\documentclass[conference]{IEEEtran}
\usepackage[none]{hyphenat}

\usepackage{cite}
\usepackage{amsmath,amssymb,amsfonts}
\usepackage{algorithmic}
\usepackage{graphicx}
\usepackage{textcomp}
\usepackage{xcolor}
\def\BibTeX{{\rm B\kern-.05em{\sc i\kern-.025em b}\kern-.08em
    T\kern-.1667em\lower.7ex\hbox{E}\kern-.125emX}}

\def\IEEEkeywordsname{Keywords}

\begin{document}

\title{Deep Learning in Medical Image Classification from MRI-based Brain Tumor Images\\

}

\author{\IEEEauthorblockN{Xiaoyi Liu}
\IEEEauthorblockA{\textit{Ira A. Fulton Schools of Engineering} \\
\textit{Arizona State University}\\
Tempe, USA \\
xliu472@asu.edu}
\and
\IEEEauthorblockN{Zhuoyue Wang}
\IEEEauthorblockA{\textit{Department of Electrical Engineering and Computer Sciences} \\
\textit{University of California, Berkeley}\\
Berkeley, USA \\
zhuoyue\_wang@berkeley.edu}
}

\maketitle

\begin{abstract}
Brain tumors are among the deadliest diseases in the world. Magnetic Resonance Imaging (MRI) is one of the most effective ways to detect brain tumors. Accurate detection of brain tumors based on MRI scans is critical, as it can potentially save many lives and facilitate better decision-making at the early stages of the disease. Within our paper, four different types of MRI-based images have been collected from the database: glioma tumor, no tumor, pituitary tumor, and meningioma tumor. Our study focuses on making predictions for brain tumor classification. Five models, including four pre-trained models (MobileNet, EfficientNet-B0, ResNet-18, and VGG16) and one new model, MobileNet-BT, have been proposed for this study. 
\end{abstract}


\section{Introduction}
Brain tumors are diseases that are difficult to cure and have a high mortality rate. Magnetic Resonance Imaging (MRI) is one of the tests needed for diagnosing brain tumors. Accurate and timely detection through MRI scans is essential for enhancing patient outcomes. Unlike Alzheimer’s disease, which uses intelligent interaction therapy, most brain tumors need surgery \cite{lin2020touch}. In our research, we aim to utilize the brain tumor MRI dataset to classify four types of brain tumors: glioma, meningioma, pituitary tumors, and the absence of tumors. Within our paper, pre-trained models, including MobileNetV2, ResNet-18, EfficientNet-B0, and VGG16, were analyzed for the brain tumor classification task. A new model, MobileNet-BT, based on MobileNetV2, was proposed.

\section{Background}

\subsection{Deep Learning}

Machine learning is heavily used nowadays, with strategies such as random forest and XGBoost \cite{lin2024text}. Deep learning, outperformed many machine learning techniques, is widely used across different industries \cite{lecun2015deep}. Many domains, including speech recognition, image classification, and object detection, have greatly improved by using deep learning technologies, achieving cutting-edge results in various applications, including brain tumor detection and segmentation\cite{lecun2015deep}. Natural Language Processing (NLP), computer vision (CV), and many other areas benefit from deep learning technologies. For natural language processing, applications such as automatic news generation \cite{peng2024automatic}, real estate transactions \cite{zhao2024utilizing}, content recommendation systems \cite{zhao_liu_2024}, financial risk detection \cite{wang2024application}, psychiatric behavior understanding \cite{han2024chain}, and ChatGPT are good examples of NLP. The large language model can be further improved by using transformers \cite{mo2024large}. Recurrent neural network (RNN) is widely used in NLP \cite{jiang2021recurrent}. For computer vision, CNN is the widely used architecture. Other areas in deep learning that could be beneficial such as in the software development process \cite{li2024utilizing}, facial expression recognition, and credit risk management for banks \cite{cheng2024research}. Even though deep learning is growing dramatically, security is worth concern since there are many Backdoor/Trojan attacks \cite{lyu2023backdoor} \cite{lyu2024task}.

In this paper, CNN models are analyzed for brain tumor classification tasks. More specifically, pre-trained models MobileNetV2, EfficientNet-B0, ResNet-18, and VGG16 will be studied utilizing the brain tumor dataset. Additionally, a new model, MobileNet-BT, based on MobileNetV2, is introduced, which achieves higher accuracy and F1-score.

\subsection{CNN}

Convolutional Neural Networks (CNN) are extensively employed in the domain of CV. The main components of CNNs are convolutional layers, pooling mechanisms, activation functions, and fully connected networks. The convolutional layers are fundamental to the CNN architecture. Learnable filters, known as kernels, are applied to images, sliding through them to detect specific features such as edges, textures, or patterns \cite{o2015introduction}. Pooling layers decrease the spatial resolution of the images. The downsampling of feature maps enables the model to concentrate on the most important aspects while reducing the risk of overfitting \cite{o2015introduction}. Non-linear activation functions enable the model to learn complex patterns. Commonly used activation functions like ReLU help mitigate the vanishing gradient problem \cite{dubey2022activation}. Fully connected layers make the final predictions. Thanks to technologies like vision model sparsification, the more complex model could be trained in CNN  \cite{jin2023visual}. CNNs are utilized in numerous CV tasks, such as image classification, object recognition, and semantic segmentation, among others. \cite{lecun2015deep}. Many computer vision technologies can be applied to medical images. Medical image reconstruction and enhancement can help reduce noise, artifacts, and other problems \cite{gong2024research}, and models like Lightweight Information Split Network can reconstruct a super-resolution model \cite{liu2024infrared}. Additionally, nodule recognition and detection can assist in identifying abnormalities at an early stage, which could save many lives \cite{yang2024applicationcomputerdeeplearning}. 

\subsection{Pituitary Tumor}
Pituitary tumors are unusual masses that form in the pituitary gland. They are very common, but fortunately, most are invariably benign, which means they are not cancer. Pituitary tumors may hypersecrete hormones, which include but are not limited to prolactin, growth hormone, and adrenocorticotropic hormone\cite{melmed2011pathogenesis}. The pituitary tumor MRI-based image can be found in Fig.1 below.

\subsection{Meningioma Tumor}

Meningiomas are the most common primary intracranial tumor. Although often perceived as entirely benign, meningiomas are frequently linked to significant health issues, including localized neurological impairments, seizures, and reduced quality of life\cite{buerki2018overview}. The meningioma tumor MRI-based image can be found in Fig.1 below.

\subsection{Glioma Tumor}

Gliomas are the most prevalent type of brain tumor. Gliomas are classified into four grades, ranging from grade I, which is benign (noncancerous), to grade IV, which is the most aggressive. Grade IV gliomas are the most aggressive and frequently occurring gliomas \cite{jiang2012origin}. The MRI-based image of the glioma tumor can be found in Fig.1 below.

\begin{figure}[ht]
    \centering
    \includegraphics[width=1\linewidth]{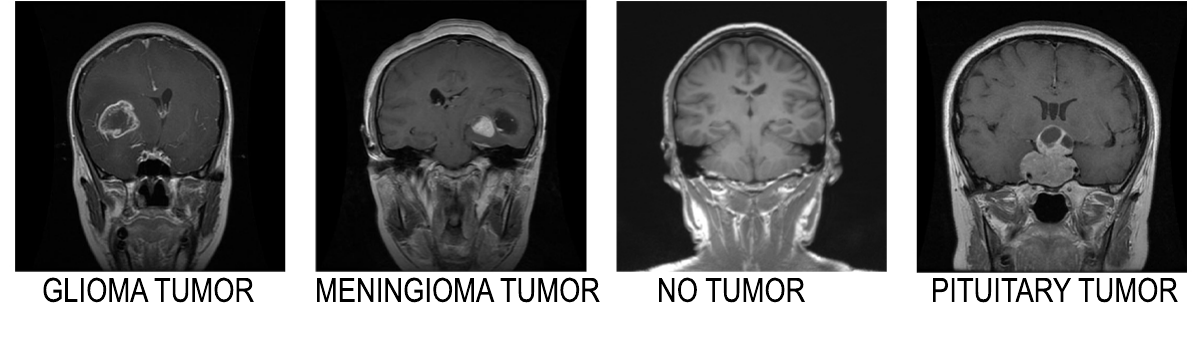}
    \caption{Brain Tumor Types}
    \label{fig: Brain Tumor Types}
\end{figure}

\subsection{Pre-trained Models}

\subsubsection{MobileNetV2}

MobileNets are designed to work efficiently, balancing efficiency and accuracy. The model is designed to work in mobile or other resource-constrained environments, reducing the required computational power while maintaining high accuracy \cite{sandler2018mobilenetv2}. For MobileNetV2, inverted residuals, linear bottlenecks, and ReLU activation functions are introduced, achieving better accuracy compared to MobileNetV1 while maintaining a relatively small model size\cite{sandler2018mobilenetv2}.

\subsubsection{EfficientNet-B0}

EfficientNet models have different versions, starting with EfficientNet-B0 and going up to B7. These versions increase in size and complexity using a new scaling technique called compound scaling. This method for EfficientNet can be described by the following formulas:

\begin{align}
\text{Depth:} & \quad d = \alpha^\phi \\
\text{Width:} & \quad w = \beta^\phi \\
\text{Resolution:} & \quad r = \gamma^\phi
\end{align}

Where:
\begin{itemize}
    \item $\alpha$, $\beta$, and $\gamma$ are the scaling coefficients for depth, width, and resolution, respectively.
    \item $\phi$ is a user-defined parameter that determines the overall scaling factor.
\end{itemize}

Using compound scaling can dramatically increase performance when compared to other scaling techniques such as width scaling, depth scaling, and resolution scaling \cite{tan2019efficientnet}.

\subsubsection{ResNet-18}

ResNet-18, introduced by Kaiming He and colleagues in 2016, is a convolutional neural network architecture. The residual blocks are the main innovation addressing vanishing/exploding gradient problems. It allows the CNN model to go deeper\cite{he2016deep}. The basic residual block used in ResNet-18 can be described by the following equation:

\begin{equation}
    y = \mathcal{F}(x, \{W_i\}) + x
\end{equation}

Where:

\begin{itemize}
    \item $x$ represents the input to the residual block.
    \item $\mathcal{F}(x, \{W_i\})$ represents the stacked convolutional layers, batch normalization, and ReLU activations.
    \item $y$ represents the output of the residual block.
\end{itemize}

\subsubsection{VGG16}

The final convolutional neural network architecture examined in this paper is VGG16. It was first proposed by Visual Geometry Group (VGG) in 2014. The architecture includes 13 convolutional layers and 3 dense layers. It achieved great performance on the ImageNet dataset at that time. The downside of VGG is having 138 million parameters, which could result in slow training time \cite{simonyan2014very}.

\begin{table*}[ht!]
\centering

\caption{Comparison of Models on Brain Tumor Classification}
\label{table:results}

\begin{tabular}{lcccccc}
\hline
\textbf{Model} & \textbf{Ave. Loss} & \textbf{Accuracy} & \textbf{Precision} & \textbf{Recall} & \textbf{F1-Score} \\
\hline
MobileNetV2 & 0.3840 & 0.8445 & 0.8498 & 0.8445 & 0.8431 \\
ResNet-18 & 0.3265 & 0.8659 & 0.8658 & 0.8659 & 0.8635 \\
EfficientNet-B0 & 0.2964 & 0.8933 & 0.8961 & 0.8933 & 0.8919 \\
VGG16 & 0.1602 & 0.9497 & 0.9495 & 0.9497 & 0.9494 \\
\textbf{MobileNet-BT} & \textbf{0.0342} & \textbf{0.9924} & \textbf{0.9924} & \textbf{0.9924} & \textbf{0.9924} \\
\hline
\end{tabular}

\end{table*}

\section{Methodology}

\subsection{Dataset}
The Brain Tumor MRI Dataset is a publicly available dataset used in this research paper \cite{msoud_nickparvar_2021}. It comprises 7023 images, with 2000 images without tumors, 1757 pituitary tumor images, 1621 glioma tumor images, and 1645 meningioma tumor images. The dataset is subsequently split into 0.8 for training, 0.1 for validation, and 0.1 for testing. In this research, supervised learning is used since all the labels are assigned to each image. When data is unlabeled, self-supervised learning strategies can be applied \cite{zhao2024optimization}.

\subsection{Image Augmentation and Preprocessing}
For training datasets, image augmentation and preprocessing are used to avoid overfitting, especially when the dataset has a relatively small number of training images. Images are randomly flipped along both horizontal and vertical axes with a 50\% chance, the images are then randomly rotated by ±10 degrees. After that, all images are resized into 224 by 224 pixels. This step is necessary for batch processing in the CNN models. Since all images from datasets are in squares, there is no additional need for cropping the images. Then, all images are converted into tensors with the range of [0, 1]. Finally, normalization is applied to the training images.

\subsection{Learning Rate and Epoch}
The learning rate started with 0.001 when training. Then, a learning rate scheduler was applied. The learning rate will be multiplied by 0.1 every 8 epochs. It will help the model to converge properly. The number of epochs is determined by the validation dataset's accuracy. If the validation accuracy does not improve for 8 successive epochs, the training stops. This will make sure no overfitting was done for the training model.

\subsection{Pretrained Models}

Four pre-trained models were tested with the dataset: MobileNetV2, ResNet-18, EfficientNet-B0, and VGG16 achieved accuracies of 0.8445, 0.8659, 0.8933, and 0.9497, respectively, with the number of parameters being 3.5 million, 11.7 million, 5.3 million, and 138 million, respectively. There are techniques such as Learning from Teaching (LoT) that could enhance generalization and potentially achieve better outcomes \cite{jin2024learning}. 

All four models were implemented using transfer learning, with all layers except the last one frozen. The accuracy and F1-score of the test dataset are good but could be better with additional operations. 

\subsection{Proposed Deep Learning Models}

Pre-trained models are typically trained on extensive datasets like ImageNet. The size of ImageNet and the dataset we use in this research differ significantly. Medical image classification is far different from generic image classification. By unfreezing more layers from the pre-trained model, we can potentially make better predictions for specific datasets and tasks. After comparing all four models, we observed that although MobileNetV2's accuracy is slightly lower compared to the other models, with only 3.5 million parameters and an accuracy of 0.8445, the training time and efficiency of using MobileNetV2 are very competitive when compared with VGG16, which uses 138 million parameters to achieve an accuracy of 0.9497.

Based on the analysis above, the MobileNet-BT model was used in the brain tumor classification task. MobileNet-BT utilized the pre-trained weights of MobileNetV2. After that, all the layers of MobileNetV2 were unfrozen. This step enables the model to capture more specific characteristics from the brain tumor dataset. Then, the final layer was replaced by a custom classifier. The fully connected layer initially incorporates a dropout layer with a probability of 20\%, which aids in avoiding overfitting. It is then followed by a linear layer with 1000 output features. An additional dropout layer with a probability of 20\% was included. Lastly, a linear layer with four output classes was added. The customized classifier allows the model to learn different levels of abstraction. The dropout layers, which randomly deactivate some neurons, help prevent the model from relying too heavily on certain neurons. The additional dense layer enables the model to capture more complex patterns from the dataset.

\subsection{Performance Measure}

When measuring the performance of different models on the test set, we used accuracy,  F1-score, precision, recall, and average loss as the metrics. All of these are good measurements for classification tasks. TP (True Positive), FP (False Positive), FN (False Negative), and TN (True Negative) are used in the equations below. All the equations can be found as follows:

\begin{equation}
\text{Accuracy} = \frac{TN + TP}{TP + FP + TN + FN}
\end{equation}

\begin{equation}
\text{Recall} = \frac{TP}{FN + TP}
\end{equation}

\begin{equation}
\text{Precision} = \frac{TP}{TP + FP}
\end{equation}

\begin{equation}
\text{F1-Score} = \frac{2 \cdot \text{Precision} \cdot \text{Recall}}{\text{Recall} + \text{Precision}}
\end{equation}

For average loss, cross-entropy was used to calculate average loss, where \( N \) is the number of classes, \( y_i \) is the true label, and \( \hat{y}_i \) is the predicted probability for class \( i \). The equation is given below:

\begin{equation}
L = -\sum_{i=1}^{N} y_i \log(\hat{y}_i)
\end{equation}

Cross-entropy was used as the criterion for all tests conducted in this paper due to its wide use in classification tasks. It effectively manages the discrepancy between the predicted probabilities and the actual outcomes. \cite{bishop2006pattern}.

\begin{figure}[!ht]
    \centering
    \includegraphics[width=1\linewidth]{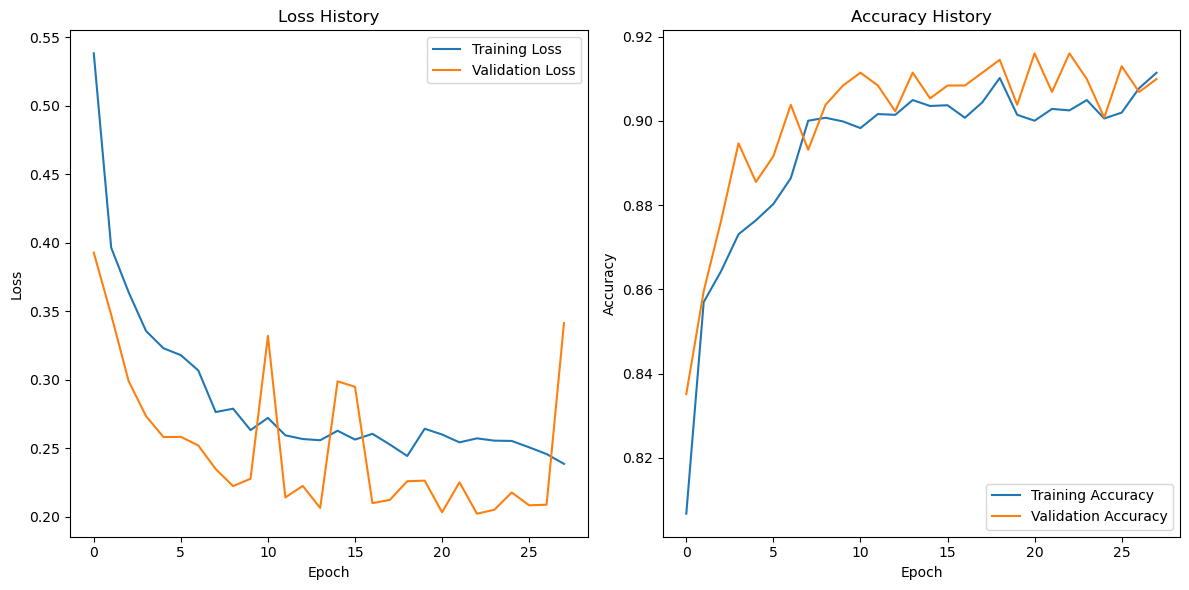}
    \caption{EfficientNetB0 Loss VS Accuracy}
    \label{fig:EfficientNetB0 Loss VS Accuracy}
\end{figure}

\begin{figure}[!ht]
    \centering
    \includegraphics[width=1\linewidth]{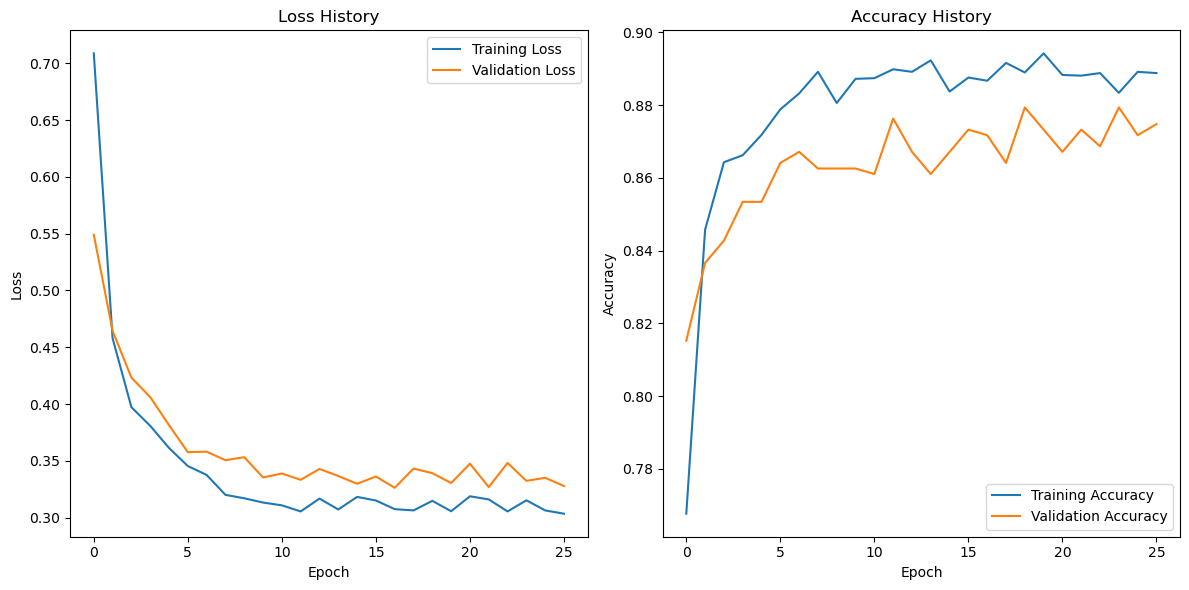}
    \caption{MobileNetV2 Loss VS Accuracy}
    \label{fig: MobileNetV2 Loss VS Accuracy}
\end{figure}

\begin{figure}[!ht]
    \centering
    \includegraphics[width=1\linewidth]{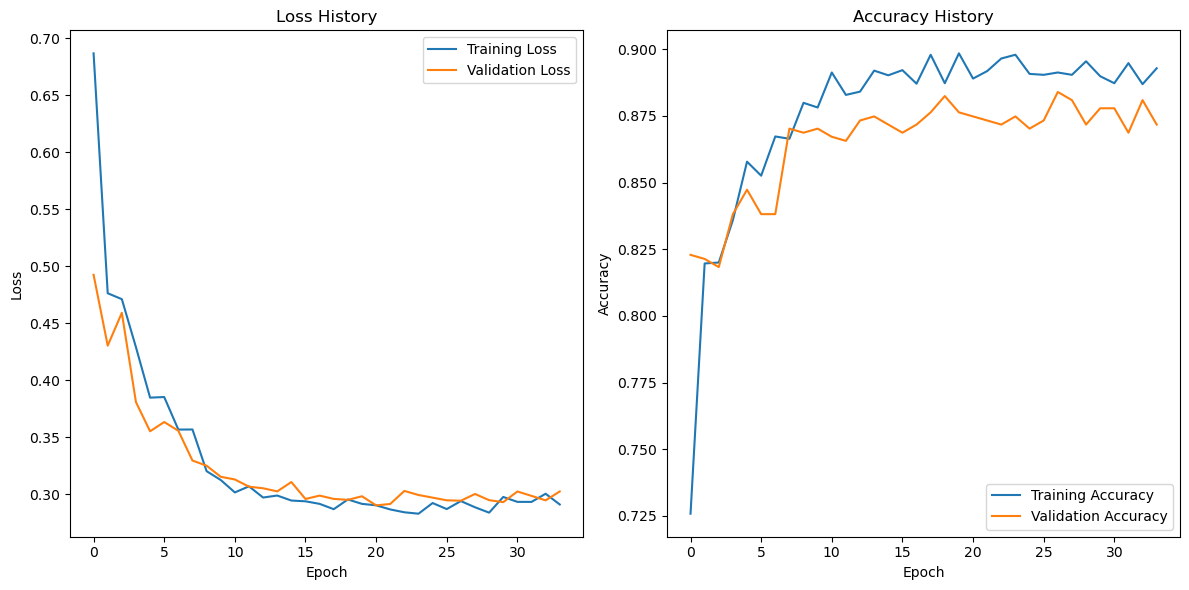}
    \caption{Resnet18 Loss VS Accuracy}
    \label{fig: Resnet18 Loss VS Accuracy}
\end{figure}

\begin{figure}[!ht]
    \centering
    \includegraphics[width=1\linewidth]{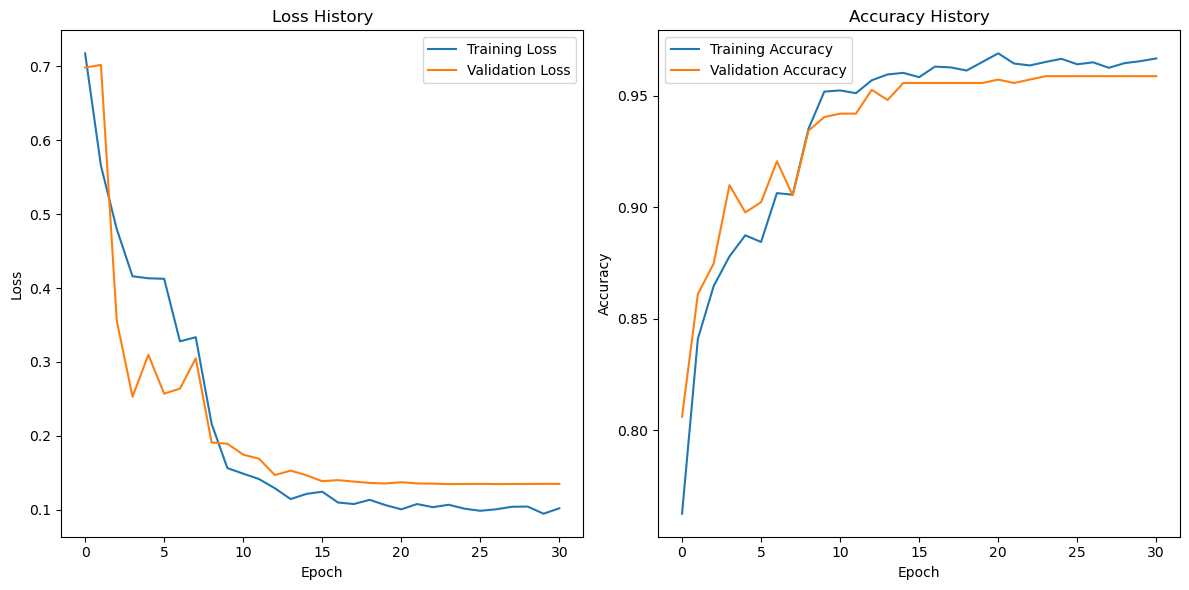}
    \caption{VGG16 Loss VS Accuracy}
    \label{fig: VGG16 Loss VS Accuracy}
\end{figure}

\begin{figure}[!ht]
    \centering
    \includegraphics[width=1\linewidth]{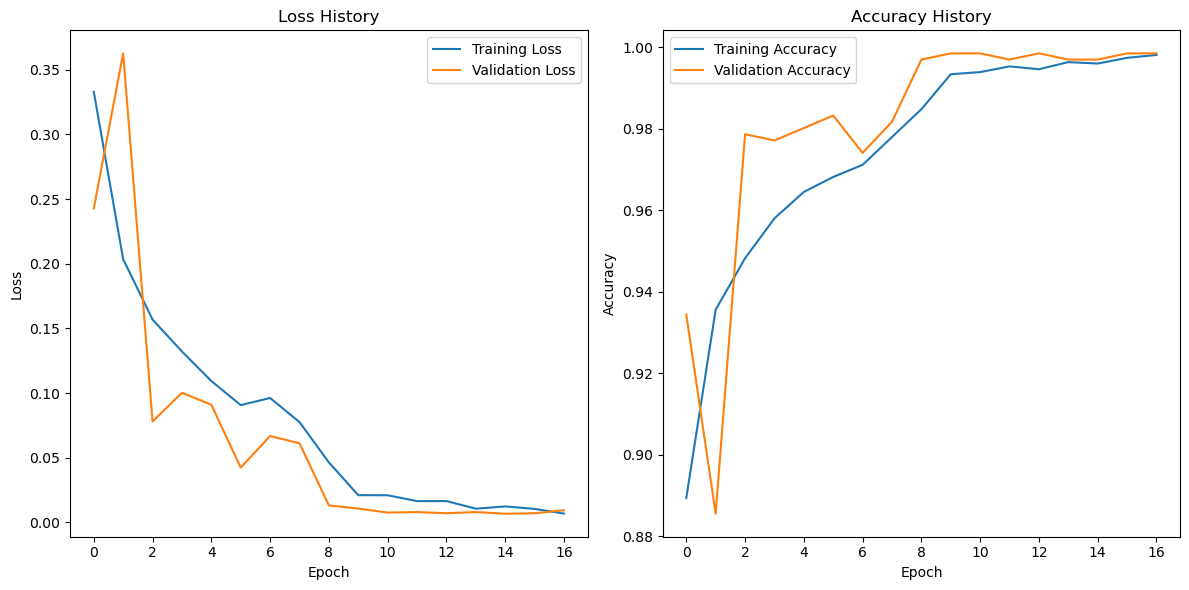}
    \caption{MobileNet-BT Loss VS Accuracy}
    \label{fig: MobileNet-BT Loss VS Accuracy}
\end{figure}

\section{Evaluation and Discussion of Results}

Different metrics are applied when comparing the performance of different models for the brain tumor classification task. For pre-trained models, MobileNetV2, ResNet-18, EfficientNet-B0, and VGG16 achieved accuracies of 0.8445, 0.8659, 0.8933, and 0.9497, respectively, and F1-scores of 0.8431, 0.8635, 0.8919, and 0.9494 in that order. Compared to these pre-trained models, the customized model MobileNet-BT achieved much higher scores in all metrics, with 0.9924 for the accuracy and 0.9924 for the F1-score. More information on the results can be found in Table I above. 

The revision of unfreezing all layers and modifying the classifier significantly improved the performance of medical image classification. While pre-trained models perform well in medical image classification, significant differences in dataset size and image types between generic and medical images require adjustments. Customizing the architecture and unfreezing layers of pre-trained models can dramatically enhance performance in medical image classification. MobileNet-BT attained an accuracy of 99.2\%, in contrast to the plain MobileNetV2 model with an accuracy of 84.5\%. This indicates the importance of further customizing and fine-tuning existing models to achieve significantly higher performance in medical image classification. 

Fig.2, 3, 4, 5, and 6 illustrate the loss and accuracy over epochs. MobileNet-BT achieved its best model at the 10th epoch, while others achieved their best models around or after the 20th epoch. This also indicates that the new model, MobileNet-BT, not only improves accuracy and other metrics such as F1-score but also makes predictions more efficiently.

\section{Conclusion}

Four pre-trained models were tested on the brain tumor dataset, resulting in accuracies of 0.8445, 0.8659, 0.8933, and 0.9497, respectively. The new model, MobileNet-BT, which is based on MobileNetV2, achieved an accuracy of 0.9924. Meanwhile, fewer epochs are required to train the new model compared to the pre-trained models. This model mainly focuses on medical image classification tasks and achieves much higher accuracy in the brain tumor classification task than other pre-trained models. Even though the pre-trained models were trained with much larger datasets for specific tasks, especially medical images, a revised model and customized architecture are required to achieve better outcomes.

\bibliography{reference.bib}
\bibliographystyle{IEEEtran}

\vspace{12pt}
\end{document}